\documentclass[10pt,twocolumn,letterpaper]{article}

\usepackage{iccv}
\usepackage{times}
\usepackage{epsfig}
\usepackage{graphicx}
\usepackage{amsmath}
\usepackage{amssymb}
\usepackage{subcaption}

\usepackage{fancyhdr}

\usepackage[pagebackref=true,breaklinks=true,letterpaper=true,colorlinks,bookmarks=false]{hyperref}

\iccvfinalcopy 

\def\iccvPaperID{2} 

\DeclareMathOperator*{\argmin}{argmin}

\fancypagestyle{plain}{
\fancyhf{} 
\fancyfoot[L]{
\copyright2017 IEEE\@. Personal use of this material is permitted. Permission from IEEE must be obtained for all other uses, in any current or future media, including reprinting /republishing this material for advertising or promotional purposes, creating new collective works, for resale or redistribution to servers or lists, or reuse of any copyrighted component of this work in other works. \hfill}
\fancyfoot[C]{}
\fancyfoot[R]{}

}

\begin{document}

\title{Pix2Face: Direct 3D Face Model Estimation}

\author{Daniel Crispell \hspace{2mm} Maxim Bazik\\
  Vision Systems, Inc.\\
  Providence, RI USA\\
  {\tt\small daniel.crispell, maxim.bazik@visionsystemsinc.com}
}

\maketitle

\begin{abstract}
  An efficient, fully automatic method for 3D face shape and pose estimation in unconstrained 2D imagery is presented. The proposed method jointly estimates a dense set of 3D landmarks and facial geometry using a single pass of a modified version of the popular ``U-Net'' neural network architecture.  Additionally, we propose a method for directly estimating a set of 3D Morphable Model (3DMM) parameters, using the estimated 3D landmarks and geometry as constraints in a simple linear system.  Qualitative modeling results are presented, as well as quantitative evaluation of predicted 3D face landmarks in unconstrained video sequences.
\end{abstract}


\section{Introduction}
Automatic estimation of 3D face shape and pose ``in the wild'' has many practical applications, including performance capture, biometrics, and advanced image processing (e.g.\ change of expression, lighting, or pose).
In order to solve the problem without assuming any prior information about the imaging device or head pose, existing approaches typically involve an optimization approach where camera, pose, and face shape parameters are iteratively refined from a rough initial estimate.
In contrast to existing approaches, the proposed modeling pipeline directly estimates dense 3D facial geometry from an input image using an encoder/decoder-style convolutional neural network (CNN). While dense 3D geometry is itself useful for some applications, it is often useful to have a parameterized 3D model describing the full facial geometry, including that which is not visible in the input image. Given a 3D Morphable Model (3DMM)~\cite{blanz_morphable_1999, booth_3d_2016} constructed from a set of training subjects, a straightforward method for estimating the coefficients using the estimated 3D geometry is presented.

The remainder of the paper is laid out as follows. A selection of prior relevant work is presented in Section~\ref{sec:related_work}. The details of the proposed 3D Modeling approach are described in Section~\ref{sec:3D_facial_modeling}. Using results of the proposed approach, a method for estimating anatomical facial landmarks is described in Section~\ref{sec:landmark_localization}, and quantitative results are presented. Finally, the paper is concluded in Section~\ref{sec:conclusion}.

\subsection{Contributions}
The contributions of this work are:
\begin{itemize}
    \item A direct image-to-3D estimation system based on a multi-output Convolutional Neural Network.
    \item An efficient camera model and facial structure estimation algorithm based on the result of the direct image-to-3D estimation.
    \item A method for robust 2D and 3D landmark localization using the estimated model.
\end{itemize}

\begin{figure}[t]
  \centering
  \includegraphics[width=0.5\textwidth]{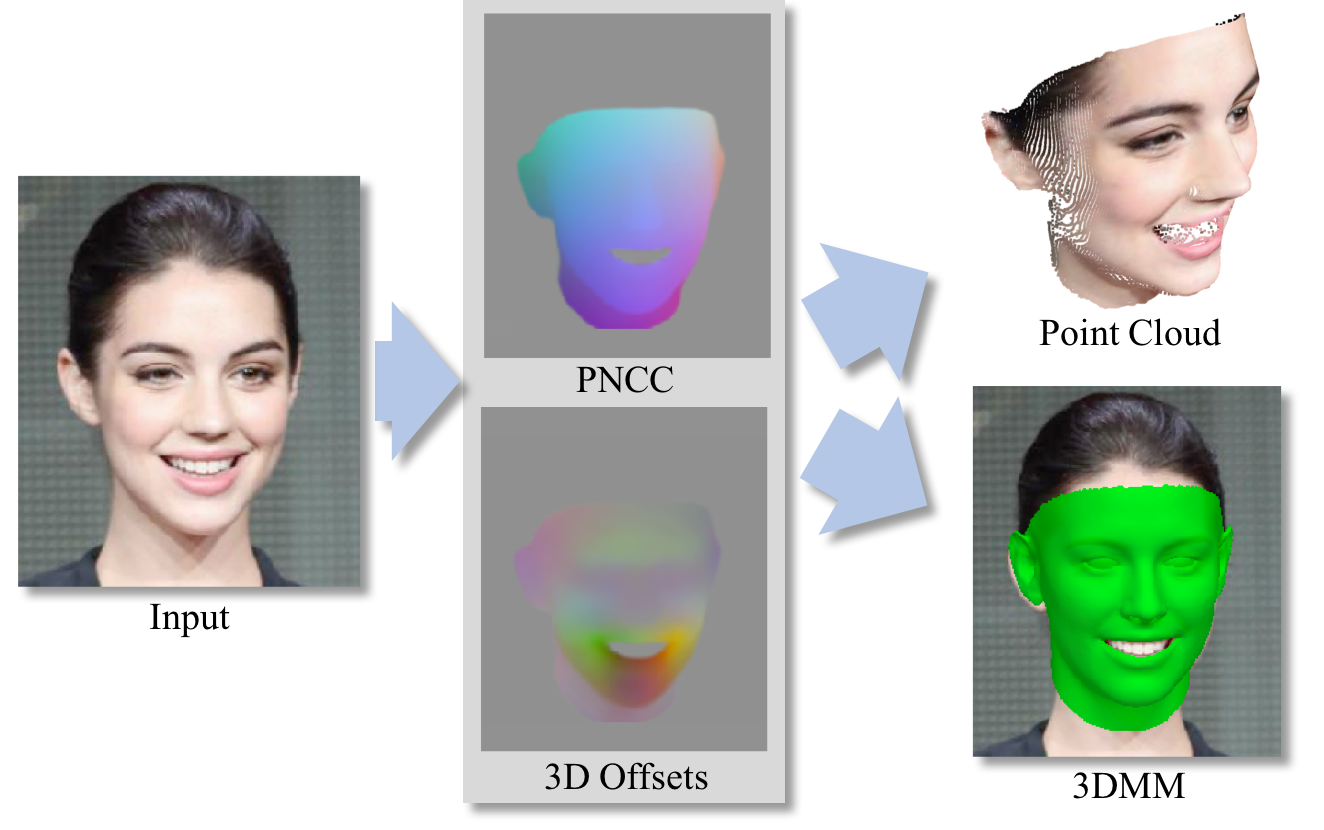}
  \caption{Overview of the proposed 3D facial modeling pipeline. From left to right: From the input image, a trained CNN predicts a PNCC (dense facial landmarks) and 3D offset image, which fully specify a 3D point cloud.  Alternatively, coefficients of a 3DMM may be estimated using the predicted 3D geometry as constraints.}
\label{fig:pix2face_teaser}
\end{figure}

\section{Related work}
\label{sec:related_work}
Many approaches to ``in the wild'' 3D face modeling leverage a parameterized model of 3D face geometry and optimize the parameter values based on 2D image-level constraints.  By far the most popular such model is the 3D Morphable Model (3DMM) of Blanz and Vetter~\cite{blanz_morphable_1999}. The 3DMM represents deviations from a canonical ``mean face'' geometry and appearance using a linear model based on principal components analysis (PCA) of 200 3D face scans.  Recently, Booth et al.~\cite{booth_3d_2016} released a larger scale 3DMM based on 10,000 face scans with improved ethnic and age diversity.

The originally proposed approach to 3DMM coefficient estimation~\cite{blanz_morphable_1999} relied on a non-linear optimization based on minimizing the differences between predicted and observed image intensities. Because of this, the method required careful initialization and was computationally expensive to optimize. More recent optimization approaches~\cite{zhu_high_2015,crispell_dataset_2016} rely on sparse 2D landmarks, which can be computed efficiently and robustly~\cite{kazemi_one_2014}. Because the detected landmarks are inherently 2D, however, these methods must account for occlusion and pose effects by incorporating an iterative optimization that alternates between updating camera, pose and geometry parameters.

In order to avoid the problem of viewpoint-dependent 2D landmarks, Zhu et al.~\cite{zhu_face_2016} introduced the projected normalized coordinate code (PNCC), which acts as a fully-3D set of dense facial landmarks.  They leverage the PNCC as an intermediate output in a cascade of convolutional neural networks, each of which produce updates to the current 3DMM coefficient estimates given the input image and a predicted PNCC, which is rendered using the current coefficient estimates at each stage. The proposed method also leverages the idea of the PNCC image, but directly estimates the PNCC (in conjunction with an image of dense 3D offsets) using a single network given only the original image as input. 3DMM coefficients are then fit (if desired) directly to the predicted 3D output.

By making some simplifying assumptions about the reflectance properties of the face, it is possible to estimate fine-scaled 3D facial geometry using shape from shading (SfS) constraints.  Although recent approaches~\cite{kemelmacher-shlizerman_face_2011,roth_adaptive_2016, trigeorgis_face_2017} have shown applicability in ``real-world'' scenarios, we avoid making assumptions about lighting, imaging, and reflectance conditions with the aim of producing as robust a system as is possible.

\section{3D Facial Modeling}
\label{sec:3D_facial_modeling}
The proposed facial modeling pipeline is composed of two stages: Direct 3D shape estimation, and 3DMM coefficient estimation.  Direct 3D shape estimation is accomplished through a convolutional neural network (CNN) trained to predict two output images: A dense landmark image, and dense set of 3D offsets from a canonical ``mean face'' baseline shape.  Given these two images, a camera model is estimated using 2D to 3D correspondences, and a set of 3DMM coefficients are solved for via a single linear system of equations.

\subsection{The Pix2face Network}
The pix2face network accepts a single standard RGB image of a cropped face as input, and produces two 3D images (a total of six 2D image planes) as output.  The first output is a dense landmark image, termed ``projected normalized coordinate code'', or PNCC, by Zhu et al.~\cite{zhu_face_2016}. The PNCC image represents the corresponding point on the ``mean face'' for each pixel of the input image by its 3D coordinate. For example, the pixel corresponding to the tip of the subject's nose should always map to a PNCC value equivalent to the 3D coordinate of the tip of the nose on the ``mean face''.  The second output is a second set of 3D offsets from the mean face, representing the deviation of the subject's face geometry (including expression) from the ``mean face'' model.  By adding the PNCC values and the offset values corresponding to a given input image, an estimated 3D point cloud is produced.  While it is feasible to train a network to directly output the 3D coordinates, it is useful to have the PNCC image and offsets as distinct entities for the purpose of registration (e.g.\ for computing a 3DMM as in Section~\ref{sec:coeff_estimation}).

\subsubsection{Network architecture}
We use a slightly modified ``U-Net''~\cite{ronneberger_unet_2015} architecture for the pix2face network. The U-Net architecture consists of a symmetric encoder and decoder pair, with skip connections between each pair of corresponding layers in the encoder and decoder. The skip connections allow the decoder network to leverage high-resolution information available in the early layers of the encoder.  Our implementation produces six planes of output values at identical resolution of the input image and is trained using an $L_1$ loss function with respect to the ground truth PNCC and offset images. In addition, we found that replacing the transposed convolution layers in the decoder with convolution plus upsampling helped alleviate small ``checker board'' artifacts~\cite{odena_deconvolution_2016} in the outputs.

\subsection{Training}
A major challenge in training the proposed network is acquiring a sufficiently large set of (image, PNCC, offset) image triplets to serve as ground truth.  We leverage the 300W and 300W-LP datasets~\cite{zhu_face_2016}, which contain approximately 126000 facial images and corresponding sets of 3DMM coefficients. The dataset is a mix of real images drawn from various publically available databases, as well as a set of semi-synthetic renderings generated by Zhu et al.~\cite{zhu_face_2016} for the purpose of increased training set pose diversity. Using the provided coefficients, we render PNCC and offset images based on the camera model and face geometry described by the coefficients.

\subsubsection{Implementation Details}
The network is then trained for a total of 60 epochs, with each epoch consisting of 100000 images in total.  Network weights are updated using the Adam optimizer~\cite{kingma_adam_2015} with a minibatch size of four images.  The learning rate parameter of Adam is initialized to 0.001, and decreases in each epoch according to a fixed schedule to a minimum value of approximately $2.1 \times 10^{-5}$.  The momentum decay rate parameters are kept fixed at values of 0.5 and 0.9, respectively.  In practice, we found that a minibatch size of 4 produced models that generalized better than those that were trained with larger minibatch sizes.  The performance appeared to be relatively stable with respect to learning rate.

In order to add variation to the training set and improve generalization, two types of data jittering are performed ``on the fly'' at training time: color and crop. Color jittering randomly perturbs the image's gamma value and color balance to add appearance variation.  Crop jittering randomly resizes and crops the training images in order to add variation to relative face placement.

\subsection{Unconstrained 3D Shape Estimation}
\begin{figure*}[t]
  \centering
  \includegraphics[width=0.95\textwidth]{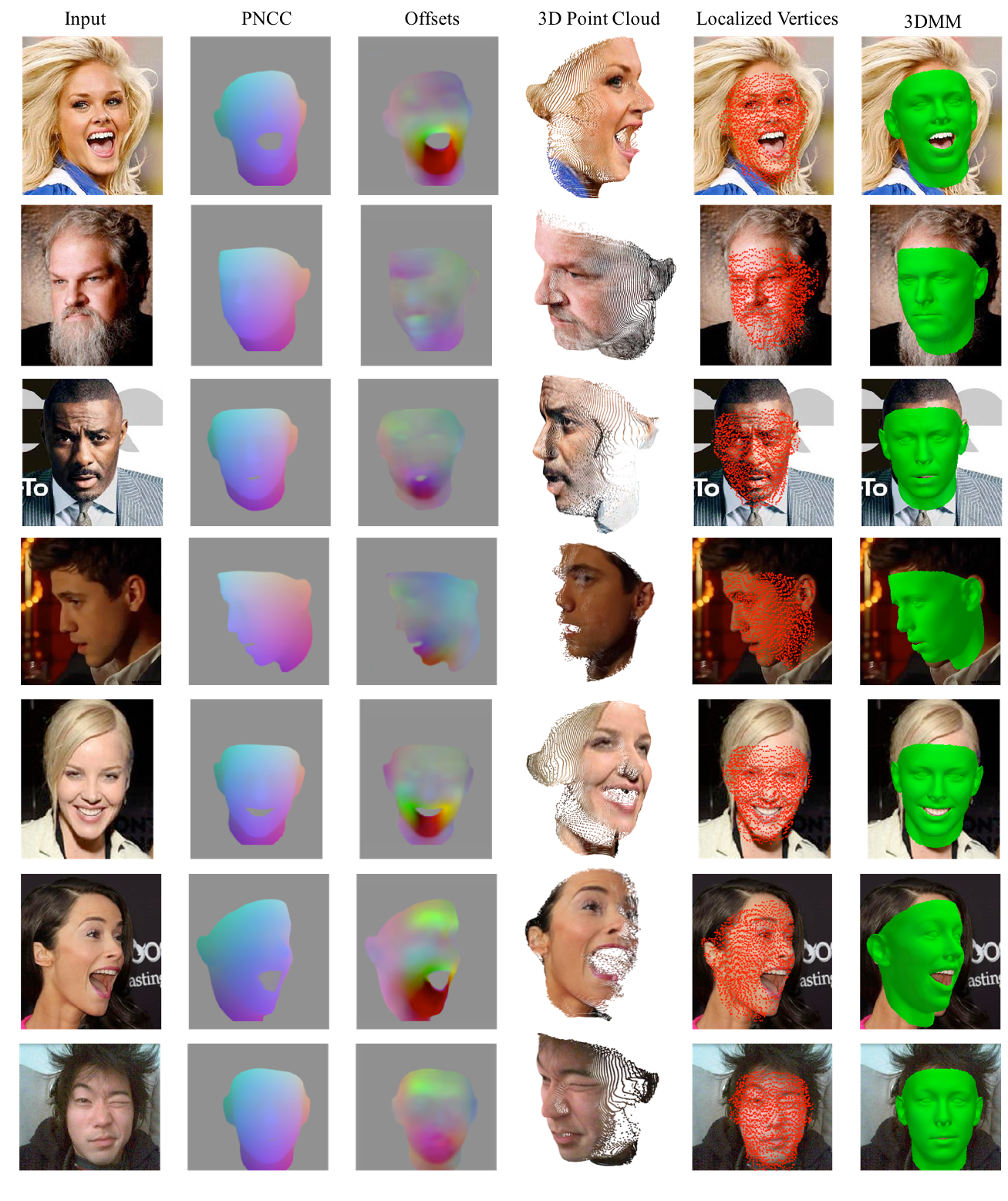}
  \caption{Qualitative results of the proposed 3D modeling pipeline on images from the VGG-Face dataset, which was not used for training. From left to right: The input image is used to predict the PNCC and Offset images. PNCC and offset $(x,y,z)$ values are shown encoded as R,G,B. By adding the PNCC and offsets, an aligned 3D point cloud is produced. Vertices of a 3DMM mesh are localized using the PNCC image (every 10th vertex shown for visualization purposes), and the corresponding 3D offsets are used to estimate the 3DMM geometry parameters, shown overlaid on top of the original image.}
\label{fig:pix2face_qual_results}
\end{figure*}

For each pixel $i$ of the input image, the trained network estimates the corresponding 3D location $\bar{x}_i$ on the ``mean face'' model, and the 3D offset $q_i$, describing the difference between the subject's facial geometry and that of the mean face at image location $i$. Adding $\bar{x}_i$ and $q_i$ results in pixel $i$'s corresponding 3D point $x_i$. For pixels corresponding to the background, the network outputs an (invalid) PNCC value near 0, and those points in the output images are ignored (shown as gray pixels in Figure~\ref{fig:pix2face_qual_results}).

The set $\{ x_i | i \in I_{\text{valid}}\}$ of estimated points for all valid pixels $I_{\text{valid}}$ comprise a 3D point cloud corresponding to the input image.  Point clouds colored by the input RGB color values are shown in Figure~\ref{fig:pix2face_qual_results}.
Note that this is accomplished directly by the network, with no explicit underlying 3D shape model (e.g. 3DMM).  In practice, however, the network implicitly relies on the 3DMM shape model due to the fact that our training data is generated using a 3DMM\@. This requirement could be relaxed if a sufficiently large dataset of dense 3D ground truth was available to train the network.

\subsection{3DMM Coefficient Estimation}
\label{sec:coeff_estimation}
Although the trained network produces an estimated 3D point for every visible pixel in the input image, it is often desirable to have a compact representation of the full 3D face geometry, including occluded surfaces.  For this purpose, a perspective camera model and shape and expression coefficients of a 3DMM are solved for, using the estimated 3D point locations as constraints.

\subsubsection{Camera estimation}
No information about the imaging device is assumed to be known a priori, rather a camera model is computed using the dense set of 2D to 3D correspondences provided by the network. Initially, each such correspondence (there are $|I_{\text{valid}}|$ in total) provides two constraints in a set of linear equations describing a 3D to 2D affine projection matrix $A$:
\begin{align}
  p_0 x_i &= u_i \\
  p_1 x_i &= v_i \nonumber
  \label{eq:affine_camera_solve}
\end{align}

where $p_0$ and $p_1$ are the first and second rows, respectively, of the affine projection matrix $A$, and pixel $i$ is located at image coordinates $(u,v)$.  While the affine projection model is a simplification of most ``in the wild'' imaging conditions, it often suffices when the distance to the subject is very large relative to the depth variation present on the face, and has the advantage that it is straightforward to directly compute.  When warranted, we relax the affine assumption and estimate parameters of a standard ``pinhole'' projective camera model, using the affine approximation to inform the initial conditions. The projective camera model is parameterized by the focal length $f$, rotation matrix $R$, and 3D translation vector $T$.  The camera is assumed to have zero skew and a principal point at the image center. The parameters are initialized using a projective approximation to $A$ (i.e. $f$ is initialized to a fixed, very large value). The parameters are then optimized using the Levenberg-Marquardt optimization algorithm to minimize projection errors of the 2D to 3D correspondences. The initial estimates derived from the affine projection matrix $A$ are typically close to optimal, ensuring that the optimization remains stable and converges quickly.

In practice, only a small subset of the $|I_{\text{valid}}|$ correspondences are needed to reliably estimate camera parameters, making both the computation of $A$ and optimization of $f$, $R$, and $T$ significantly faster to compute.

\subsubsection{Shape and Expression estimation}
The facial geometry is represented using a 3DMM, which uses principal components analysis (PCA) to represent variations of facial geometry from the ``mean face'' model.  The topology of the ``mean face'' mesh is assumed to be fixed, and the geometry is represented by the set of 3D mesh vertex positions $X$. Given a set of shape and expression PCA coefficients ($\alpha$ and $\beta$, respectively), the shape and expression PCA component matrices ($A$ and $B$, respectively), and the ``mean face'' vertex locations $\bar{X}$, $X$ is computed as follows.

\begin{equation}
  \label{eq:PCA}
  X = \bar{X} + \alpha A + \beta B
\end{equation}

In order to estimate the PCA coefficient vectors $\alpha$ and $\beta$ from the predicted dense PNCC and offset images, a set of constraints on the mesh vertex positions must be extracted. Each mesh vertex $j$ is localized to pixel coordinate $(u_j,v_j)$ by searching for the nearest PNCC pixel value $\text{PNCC}_{u,v}$ to the corresponding ``mean face'' vetex location $\bar{x_j}$.

\begin{equation}
  \label{eq:vertex_loc}
(u_j, v_j) = \argmin_{u,v} \lVert \text{PNCC}_{u,v} - \bar{x_j} \rVert^2
\end{equation}

\begin{figure}[h]
  \centering
  \includegraphics[width=0.5\textwidth]{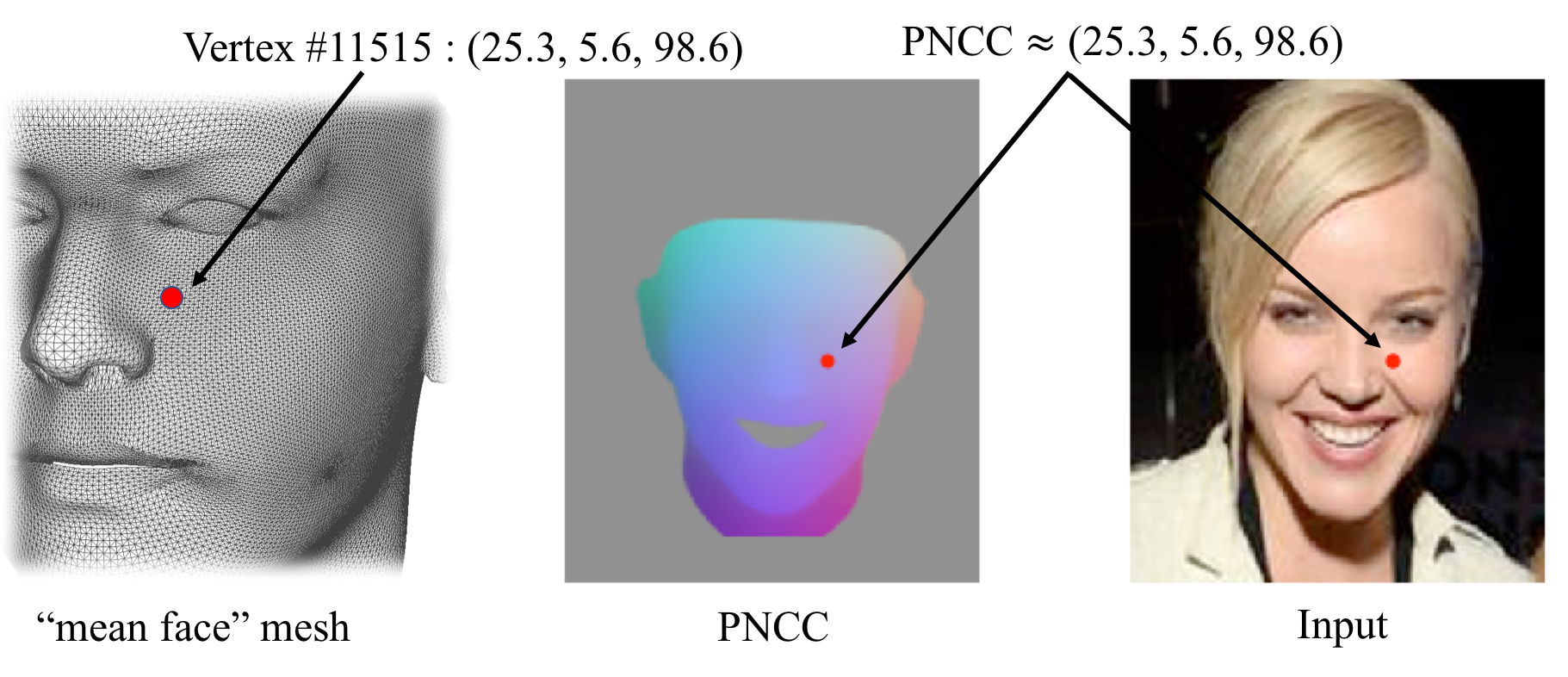}
  \caption{Mesh vertex localization: The PNCC $(x,y,z)$ pixel value best matching the coordinates of a given ``mean face'' mesh vertex are identified.}
\label{fig:vert_localization}
\end{figure}

The corresponding offset pixel value $s_j = \text{offset}_{u,v}$ is then added as a constraint to a linear system used to solve for $\alpha$ and $\beta$ consisting of all $M$ such constraints.  In general, $M$ will be significantly less than the total number of vertices since not all vertices are visible in any given image.

\begin{equation}
  \label{eq:PCA_solve}
  \begin{bmatrix}
    a_{j_0}   &   b_{j_0}   \\
    a_{j_1}   &   b_{j_1}   \\
    \vdots  &   \vdots  \\
    a_{j_M}   &   b_{j_M}   \\
\end{bmatrix}
\begin{bmatrix}
  \alpha    \\
  \beta
\end{bmatrix}
=
\begin{bmatrix}
  s_{j_0} \\
  s_{j_1} \\
  \vdots     \\
  s_{j_M}
\end{bmatrix}
\end{equation}

The per-vertex PCA component sub-matrices $a_j$ and $b_j$ are generated by extracting rows $3j$, $3j+1$, and $3j+2$ (corresponding to $x$, $y$, and $z$ vertex coordinates) from $A$ and $B$, respectively.  Tikhonov regularization using the PCA coefficient standard deviation values as weights is used to ensure the stability of the resulting PCA coefficient estimates.

If $N$ images are available of a single subject, Equation~\ref{eq:PCA_solve} can easily be adapted to solve for a single set of shape coefficients and an independent set of per-image expression coefficients $\beta_n$ as follows in Equation~\ref{eq:PCA_solve_multi}. For clarity, the individual PCA component sub-matrices $a_j$ and $b_j$ belonging to the $n$th image are concatenated into per-image block matrices $\bar{A}_n$ and $\bar{B}_n$, respectively.  Likewise, the 3D vertex offsets $s_j$ extracted from image $n$ are concatenated into $\bar{S}_n$. 

\begin{equation}
  \label{eq:PCA_solve_multi}
  \begin{bmatrix}
    \bar{A}_0   &   \bar{B}_0   & 0 & 0 & \hdots \\
    \bar{A}_1   &   0  &\bar{B}_1   & 0 & \hdots \\
    \vdots  &   & & \ddots \\
    \bar{A}_N   &  0 &  0 & \hdots & \bar{B}_N \\

\end{bmatrix}
\begin{bmatrix}
  \alpha    \\
  \beta^0 \\
  \beta^1 \\
  \vdots \\
  \beta^N
\end{bmatrix}
=
\begin{bmatrix}
  \bar{S_0} \\
  \bar{S_1} \\
  \vdots     \\
  \bar{S_N}
\end{bmatrix}
\end{equation}

Formulating the multi-image shape and expression estimation problem as a single joint optimization enforces the restriction that variation in face geometry can arise only from expression changes.  It also allows constraints from multiple viewpoints to be considered, potentially resolving ambiguities present when only a single view is available.

\section{Landmark Localization}
\label{sec:landmark_localization}
One useful application of the estimated 3D models is the localization of sparse anatomical landmarks on facial images.
We propose a simple, robust approach in which each desired landmark is associated with one of the approximately 50,000 mesh vertices of the 3DMM face model.
Each landmark is treated independently, and the optimal vertex is found as follows.
For each frame of a training set containing ground truth landmark locations, parameters of a 3DMM are estimated as described in Section~\ref{sec:coeff_estimation}.
Each mesh vertex is then projected into each image using the estimated camera parameters, and the vertex that minimizes the average reprojection error across all frames of the training set is selected.  At test time, the 3DMM parameters are estimated for an input image, and the predicted 3D mesh vertex positions are simply projected into the image using the estimated camera model. Incorporating the 3DMM allows the positions of both visible and occluded landmarks to be predicted accurately.

\begin{figure*}[h]
  \centering
  \includegraphics[width=.95\textwidth]{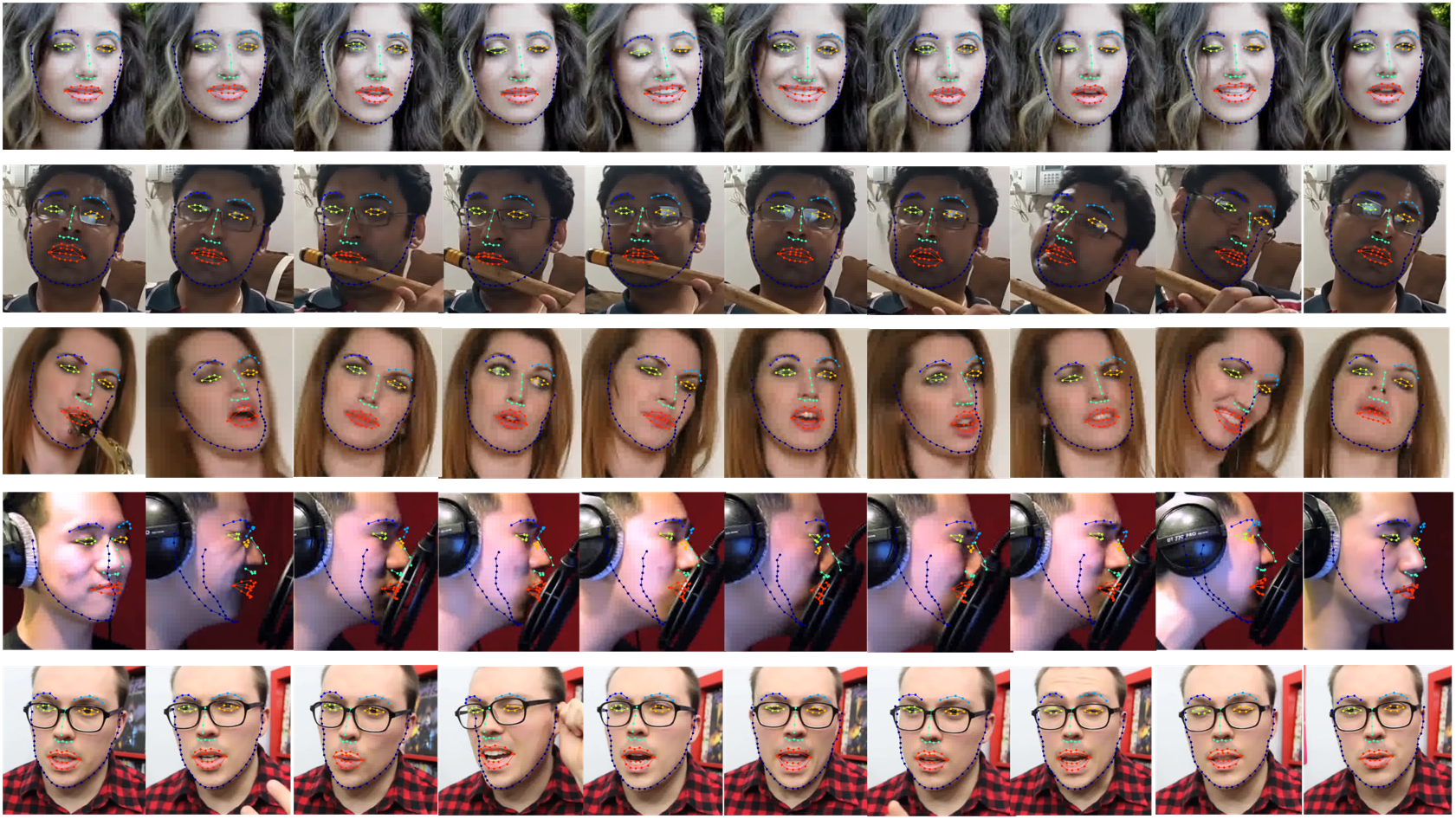}
  \caption{Representative results on individual frames from the ibug challenge test set, which contains large variations in pose, lighting, and occlusion.}
\label{fig:challenge_images}
\end{figure*}

\subsection{3D Face Tracking Challenge}
As part of the 1st 3D Face Tracking in-the-wild Competition~\cite{menpo_challenge_2017}, the proposed method was used to estimate sparse landmarks on 30 short video clips. This was accomplished using face detection, 3D model estimation, landmark interpolation, and finally smoothing.

For each image, a face bounding box is estimated using a max-margin object detector~\cite{king_max-margin_2015} as implemented in the publicly available dlib~\cite{king_dlib_2009} computer vision and machine learning library. When multiple faces are detected, the bounding box locations before and after the frame are used to determine the correct location of the subject. When the detector fails to locate any faces a bounding box is linearly interpolated from the two bounding boxes temporally adjacent to the frame.
The video frames are cropped according to the face bounding boxes and input to the pix2face network, producing PNCC and 3D offset images which are then used to estimate the face shape and camera parameters.
The 3D position of each landmark vertex is recorded, and the 2D position is obtained by projection using the estimated camera model.
The 2D landmark locations are smoothed using a moving average over three frames, which shows slight accuracy improvement on training data.

\begin{figure*}[h]
   \centering
   \begin{subfigure}[h]{.4\textwidth}
       \caption{Mean difference in 2d landmark locations normalized by interocular distance in pixels.}
\label{fig:challenge_results_2d}
       \includegraphics[width=\textwidth]{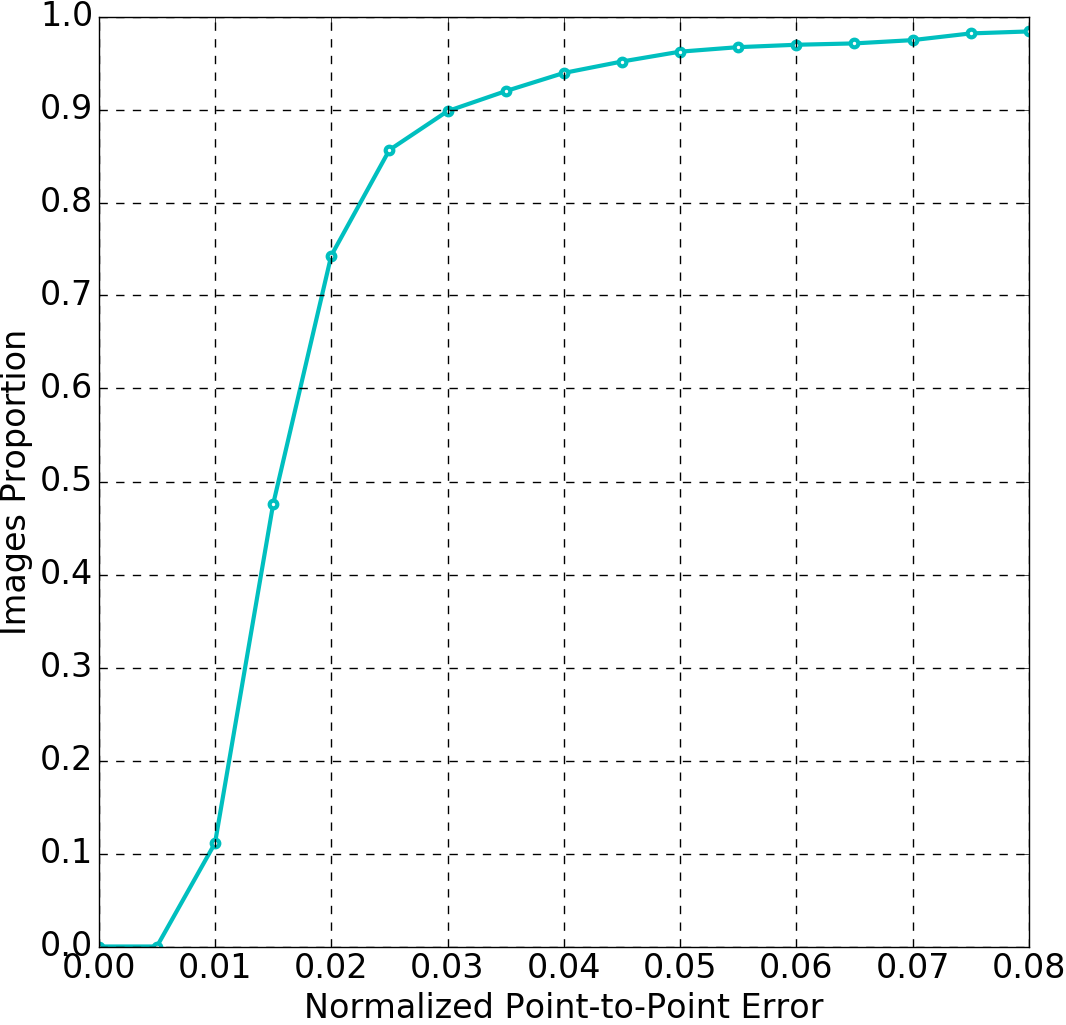}
   \end{subfigure}
   \hspace{10mm}
   \begin{subfigure}[h]{.4\textwidth}
       \caption{Mean difference in 3d landmark locations.}
\label{fig:challenge_results_3d}
       \includegraphics[width=\textwidth]{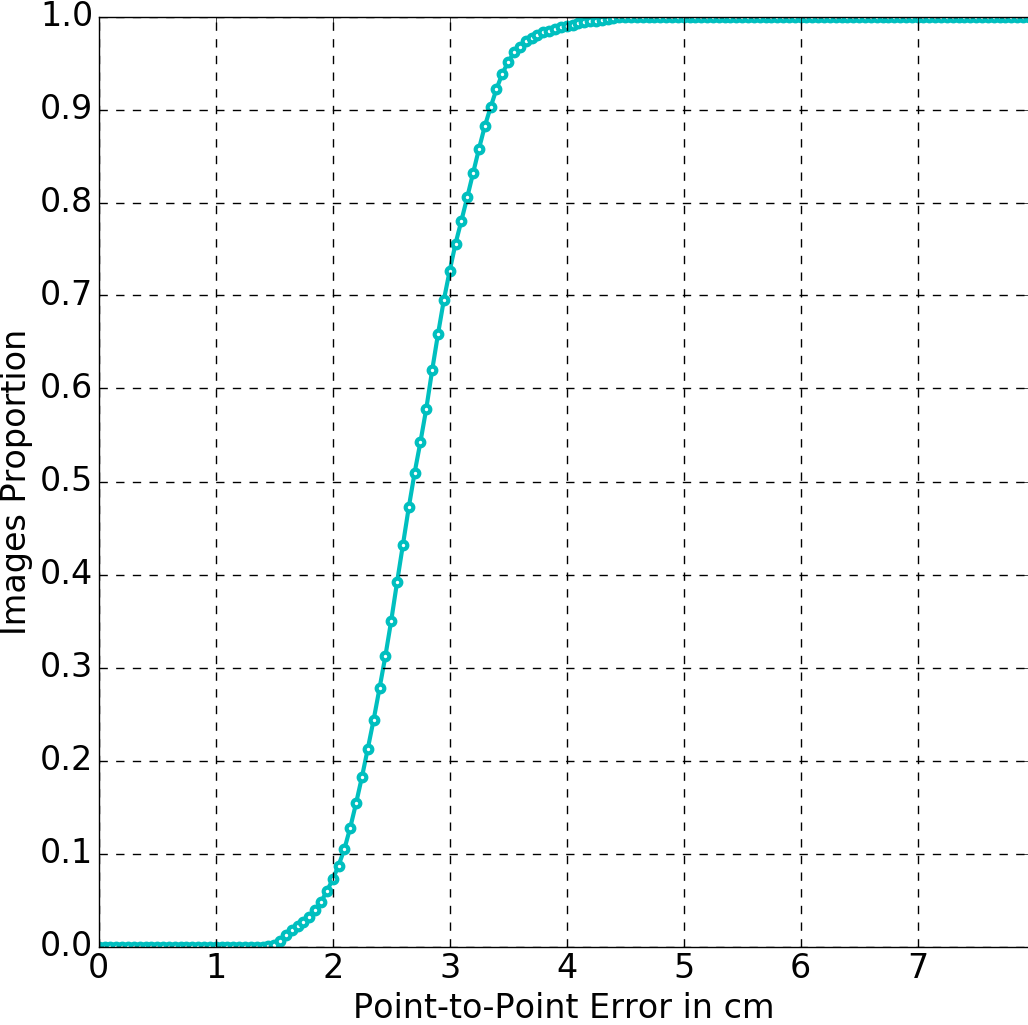}
   \end{subfigure}

  \caption{Face tracking results using the proposed method, as evaluated by the ``3D Face Tracking in-the-wild Competition''~\cite{menpo_challenge_2017} organizers.}
\label{fig:challenge_results}
\end{figure*}

Representative results from the challenge set are shown in Figure~\ref{fig:challenge_images}, and quantitative results on the test set are presented in Figure~\ref{fig:challenge_results}. The graph~\ref{fig:challenge_results_2d} reported the mean point to point error of 2d landmarks in the image. Error is measured in pixels then normalized by the interocular distance. The graph~\ref{fig:challenge_results_3d} reports the distance, measured in 3D, between estimated landmark positions and those provided by the challenge organizers.

\section{Conclusion}
\label{sec:conclusion}
As shown in Figures~\ref{fig:challenge_images}~and~\ref{fig:challenge_results}, the proposed 3D face modeling pipeline is capable of operating fully-automatically in a wide range of challenging ``in the wild'' conditions. The proposed network directly estimates registered, dense, 3D face geometry, unconstrained by any parameterized models. Future work involves further capitalizing on this capability by introducing training data which itself is not constrained by any parameterized 3D model such as registered 3D facial scans.  Additionally, a simple method for estimating 3DMM coefficients given the unconstrained 3D geometry estimates is presented.  Unlike existing approaches, the shape coefficients are solved for directly, and do not require iterative refinement.  Future work involves leveraging a 3DMM trained with more ethnic and age diversity such as that presented by Booth et al.~\cite{booth_3d_2016}.  Finally, a simple extension to the shape estimation pipeline allowing 3D and 2D anatomical facial landmarks to be estimated is presented.  Evaluation of the proposed landmark prediction pipeline on the 2017 Facial Landmark Tracking Challenge are presented.

\section*{Acknowledgement}
This research is based upon work supported by the Office of the Director of National Intelligence (ODNI), Intelligence Advanced Research Projects Activity (IARPA) under contract number 2014--14071600010.
The views and conclusions contained herein are those of the authors and should not be interpreted as necessarily representing the official policies or endorsements, either expressed or implied, of ODNI, IARPA, or the U.S. Government.
The U.S. Government is authorized to reproduce and distribute reprints for Governmental purpose notwithstanding any copyright annotation thereon.

\nocite{booth_3d_2017}

{\small
\bibliographystyle{ieee}
\bibliography{pix2face}

\begin{thebibliography}{10}\itemsep=-1pt

\bibitem{blanz_morphable_1999}
V.~Blanz and T.~Vetter.
\newblock A morphable model for the synthesis of 3d faces.
\newblock In {\em Proceedings of the 26th annual conference on {Computer}
  graphics and interactive techniques}, pages 187--194. ACM
  Press/Addison-Wesley Publishing Co., 1999.

\bibitem{booth_3d_2017}
J.~Booth, E.~Antonakos, S.~Ploumpis, G.~Trigeorgis, Y.~Panagakis, and
  S.~Zafeiriou.
\newblock 3d {Face} {Morphable} {Models} "{In}-the-{Wild}".
\newblock In {\em Proceedings of the 2017 {IEEE} {Conference} on {Computer}
  {Vision} and {Pattern} {Recognition}}, {CVPR} '17, 2017.
\newblock arXiv: 1701.05360.

\bibitem{booth_3d_2016}
J.~Booth, A.~Roussos, S.~Zafeiriou, A.~Ponniahy, and D.~Dunaway.
\newblock A 3d {Morphable} {Model} {Learnt} from 10,000 {Faces}.
\newblock In {\em 2016 {IEEE} {Conference} on {Computer} {Vision} and {Pattern}
  {Recognition} ({CVPR})}, pages 5543--5552, June 2016.

\bibitem{crispell_dataset_2016}
D.~Crispell, O.~Biris, N.~Crosswhite, J.~Byrne, and J.~L. Mundy.
\newblock Dataset {Augmentation} for {Pose} and {Lighting} {Invariant} {Face}
  {Recognition}.
\newblock In {\em 2016 IEEE Applied Imagery Pattern Recognition Workshop}.

\bibitem{kazemi_one_2014}
V.~Kazemi and J.~Sullivan.
\newblock One {Millisecond} {Face} {Alignment} with an {Ensemble} of
  {Regression} {Trees}.
\newblock In {\em Proceedings of the 2014 {IEEE} {Conference} on {Computer}
  {Vision} and {Pattern} {Recognition}}, {CVPR} '14, pages 1867--1874,
  Washington, DC, USA, 2014. IEEE Computer Society.

\bibitem{kemelmacher-shlizerman_face_2011}
I.~Kemelmacher-Shlizerman and S.~M. Seitz.
\newblock Face reconstruction in the wild.
\newblock In {\em Computer {Vision} ({ICCV}), 2011 {IEEE} {International}
  {Conference} on}, pages 1746--1753. IEEE, 2011.

\bibitem{king_dlib_2009}
D.~E. King.
\newblock Dlib-ml: {A} {Machine} {Learning} {Toolkit}.
\newblock {\em Journal of Machine Learning Research}, 10(Jul):1755--1758, 2009.

\bibitem{king_max-margin_2015}
D.~E. King.
\newblock Max-{Margin} {Object} {Detection}.
\newblock {\em arXiv:1502.00046 [cs]}, Jan. 2015.
\newblock arXiv: 1502.00046.

\bibitem{kingma_adam_2015}
D.~P. Kingma and J.~Ba.
\newblock Adam: {A} {Method} for {Stochastic} {Optimization}.
\newblock In {\em 3rd International {Conference} for {Learning}
  {Representations} ({ICLR}) 2015}, 2015.
\newblock arXiv: 1412.6980.

\bibitem{odena_deconvolution_2016}
A.~Odena, V.~Dumoulin, and C.~Olah.
\newblock Deconvolution and {Checkerboard} {Artifacts}.
\newblock {\em Distill}, 1(10):e3, Oct. 2016.

\bibitem{ronneberger_unet_2015}
O.~Ronneberger, P.~Fischer, and T.~Brox.
\newblock U-{Net}: {Convolutional} {Networks} for {Biomedical} {Image}
  {Segmentation}.
\newblock In {\em Medical {Image} {Computing} and {Computer}-{Assisted}
  {Intervention} – {MICCAI} 2015}, Lecture {Notes} in {Computer} {Science},
  pages 234--241. Springer, Cham, Oct. 2015.

\bibitem{roth_adaptive_2016}
J.~Roth, Y.~Tong, and X.~Liu.
\newblock Adaptive 3d face reconstruction from unconstrained photo collections.
\newblock In {\em 2016 {IEEE} {Conference} on {Computer} {Vision} and {Pattern}
  {Recognition} ({CVPR})}, 2016.

\bibitem{trigeorgis_face_2017}
G.~Trigeorgis, P.~Snape, I.~Kokkinos, and S.~Zafeiriou.
\newblock Face {Normals} "in-the-wild" using {Fully} {Convolutional}
  {Networks}.
\newblock In {\em 2017 {IEEE} {Conference} on {Computer} {Vision} and {Pattern}
  {Recognition} ({CVPR})}, pages 38--47, 2017.

\bibitem{menpo_challenge_2017}
S.~Zafeiriou, E.~Ververas, G.~Chrysos, G.~Trigeorgis, J.~Deng, and A.~Roussos.
\newblock The 3d {Menpo} {Facial} {Landmark} {Tracking} {Challenge}.
\newblock In {\em {ICCV} 3D {Menpo} {Facial} {Landmark} {Tracking} {Challenge}
  {Workshop}}, 2017.

\bibitem{zhu_face_2016}
X.~Zhu, Z.~Lei, X.~Liu, H.~Shi, and S.~Z. Li.
\newblock Face {Alignment} {Across} {Large} {Poses}: {A} 3d {Solution}.
\newblock In {\em 2016 {IEEE} {Conference} on {Computer} {Vision} and {Pattern}
  {Recognition} ({CVPR})}, pages 146--155, June 2016.

\bibitem{zhu_high_2015}
X.~Zhu, Z.~Lei, J.~Yan, D.~Yi, and S.~Z. Li.
\newblock High-fidelity {Pose} and {Expression} {Normalization} for face
  recognition in the wild.
\newblock In {\em 2015 {IEEE} {Conference} on {Computer} {Vision} and {Pattern}
  {Recognition} ({CVPR})}, pages 787--796, June 2015.

\end{thebibliography}
}
\end{document}